# A Compact, Hierarchically Optimal Q-function Decomposition

**Bhaskara Marthi, Stuart Russell**
Department of Computer Science
University of California, Berkeley
Berkeley, CA 94720

**David Andre**
BodyMedia Inc.
Pittsburgh, PA 15222

## Abstract

Previous work in hierarchical reinforcement learning has faced a dilemma: either ignore the values of different possible exit states from a subroutine, thereby risking suboptimal behavior, or represent those values explicitly thereby incurring a possibly large representation cost because exit values refer to nonlocal aspects of the world (i.e., all subsequent rewards). This paper shows that, in many cases, one can avoid both of these problems. The solution is based on recursively decomposing the exit value function in terms of Q-functions at higher levels of the hierarchy. This leads to an intuitively appealing runtime architecture in which a parent subroutine passes to its child a value function on the exit states and the child reasons about how its choices affect the exit value. We also identify structural conditions on the value function and transition distributions that allow much more concise representations of exit state distributions, leading to further state abstraction. In essence, the only variables whose exit values need be considered are those that the parent cares about and the child affects. We demonstrate the utility of our algorithms on a series of increasingly complex environments.

## 1 Introduction

Hierarchical reinforcement learning (HRL) aims to speed up RL by providing prior knowledge in the form of hierarchical constraints on policies [Parr and Russell, 1997; Dietterich, 2000]. In this paper, we consider policies constrained to follow a *partial program* [Andre and Russell, 2002] whose *choice points* designate places where the policy is unspecified. A learning agent equipped with such a partial program aims to find the best possible *completion*, given a set of experiences running the program in an MDP.

The fundamental result of HRL is that the combination of a partial program and an MDP yields a new semi-Markov decision process (SMDP) whose states $\omega$ are *joint* states—pairs of environment states and program states. HRL algorithms solve this SMDP; here, we consider algorithms that learn the function $Q(\omega, u)$, i.e., the expected sum of rewards when taking action $u$ in joint state $\omega$ and acting optimally thereafter. (Note that $u$ may be a temporally extended subroutine invocation as well as a primitive action.)

A principal advantage of HRL over "flat" methods is that imposing a hierarchical structure on behavior exposes structure in the value function of the underlying MDP. Roughly speaking, the complete reward sequence for any run of a policy in the MDP can be divided into subsequences, each associated with execution of a particular subroutine; this allows the global value function for the hierarchical policy to be represented as a sum of value function components, each (in the ideal case) defined over only a small set of state variables.

Two standard forms of $Q$-decomposition are known. MAXQ [Dietterich, 2000] approximates $Q(\omega, u)$ as a sum of $Q_r(\omega, u)$, the expected sum of rewards obtained while $u$ is executing, and $Q_c(\omega, u)$, the expected sum of rewards after $u$ until the current subroutine completes. [1] The ALisp decomposition [Andre and Russell, 2002] includes a third term, $Q_e(\omega, u)$, to cover rewards obtained *after* the subroutine exits. These two decompositions lead to two different notions of optimal completion. MAXQ yields *recursive optimality*—i.e., the policy within each subroutine is optimized *ignoring the calling context*. ALisp yields *hierarchical optimality*—i.e., optimality among all policies consistent with the partial program.

Recursively optimal policies may be worse than hierarchically optimal ones if context is relevant, because they ignore the exit values $Q_e$. On the other hand, ALisp may need more samples to learn $Q_e$ accurately. It can also be argued that, whereas $Q_r$ and $Q_c$ represent sums of rewards that are *local* to a given subroutine, and hence likely to be well approximated by a low-dimensional function, $Q_e$ represents a non-local sum of rewards and may, therefore, be very difficult to learn. Thus, we seem to have a dilemma: ignore $Q_e$ and risk seriously suboptimal behavior, or include $Q_e$ and incur a potentially high learning cost.

This paper shows that, in many cases, one can avoid both of these problems. The solution, described in Section 4, is based on recursively decomposing the exit value function in terms of Q-functions at higher levels of the hierarchy. New algorithms for learning and execution based on this decomposition are presented. In Section 5, we show how the idea of *additive irrelevance* can be used to dramatically further simplify the representation based on conditional independencies that may hold in the domain. Finally, in Section 6, we describe a more general simplification that depends on the size of the "interface" between a subroutine and its context. The utility of each successive decomposition is demonstrated experimentally.

[1] MAXQ also allows a "pseudoreward" function for exit values, but this must be specified in advance by the programmer.

## 2 Main example

In this section, we describe our main illustrative example. We begin with a very simple version of the environment, which will be made more complex over the course of the paper, as we introduce the various components of our approach. We use Markov decision processes as our modelling framework, and further assume that the state is represented as an instantiation to a set $\mathcal{V}$ of *state variables*. The example is based on one from [Dietterich, 2000].

**Example 1.** The taxi MDP models a taxi moving around and serving passengers on a grid. States in the MDP consist of the taxi's position and a set of passengers, each having a source, destination, generosity level, and status (`at-source` or `in-taxi`). The actions available to the taxi are: moves in the four cardinal directions, which may fail with some probability; `pickup`, which picks up the passenger at the taxi's current position, assuming the taxi doesn't already have a passenger; and `dropoff`, which drops off the passenger who is currently in the taxi, assuming the taxi is at her destination. Upon a successful `dropoff` action, with some probability the episode terminates, and otherwise, the set of passengers is reset [2] using some distribution over each passenger's source, destination, and generosity (but the taxi stays at its current position). A fixed cost is charged per time step, and a reward, which depends only on the passenger's generosity, is given after a successful dropoff.

Even for this simple domain, the optimal policy is not completely clear. For example, deciding on a passenger requires trading off the time and the expected reward for serving this passenger against how centrally placed the taxi will be after serving the passenger.

## 3 Alisp and hierarchical RL

ALisp is a language for specifying constraints on policies using "partial programs". We provide a brief overview of ALisp here. A more complete description can be found in the literature [Andre and Russell, 2002; Andre, 2003].

Figure 1 shows an example ALisp (partial) program for the environment in Example 1. The top-level function `taxi-main` says that the taxi must repeatedly serve a passenger until the episode terminates. Serving a passenger consists of choosing the passenger, then doing a `get-pass` followed by a `put-pass`. The `get-pass` subroutine consists of navigating to the passenger's source, and doing a `pickup` action in the environment. Similarly, `put-pass` consists of navigating to the passenger's destination and doing a `dropoff`. The `nav` subroutine involves repeatedly choosing a direction to move in until the goal is reached. The choices of which passenger to serve and which directions to move in while navigating are not specified by the program, but are left open using the `with-choice` command. This means that the learning algorithm has to learn a *completion* that specifies how to make these choices optimally as a function of state.

```
(defun taxi-main ()
  (loop until (at-terminal-state) do
    (call SP serve)))

(defun serve ()
  (call GP get-pass
      (choose-arg (waiting-passengers)))
  (call PP put-pass (pass-being-served)))

(defun get-pass (p)
  (call NS nav (pass-src p))
  (action PA 'pickup))

(defun put-pass (p)
  (call ND nav (pass-dest p))
  (action DA 'dropoff))

(defun nav (loc)
  (loop until (equal loc (taxi-pos))
    do (with-choice NC (d '(N S E W))
         (action NM d))))
```

Figure 1: ALisp program for basic taxi domain. ALisp-specific statements are in bold. The second argument to `call`, `with-choice`, and `action` statements is just a label. The functions `waiting-passengers`, `pass-being-served`, `pass-src`, `pass-dest`, and `taxi-pos` are not shown here; they simply extract the required components from the environment state (which is obtained using the ALisp command **`get-state`**).

Consider a partial program being executed in an environment. At any point, let $s$ denote the state of the environment, and $\theta$ denote the *machine state*, i.e., the program counter, memory, and runtime stack. Define the *joint state* $\omega = (s, \theta)$. Let $\Omega$ be the set of joint states in which the program counter is at a choice (`with-choice` or `choose-arg`) statement. Formally, a completion is a function $\pi$ on $\Omega$ such that $\pi(\omega)$ ranges over the choices available at $\omega$. A basic theorem of hierarchical RL [Parr and Russell, 1997; Precup and Sutton, 1998; Dietterich, 2000] states that given an MDP and partial program, there is an equivalent SMDP such that completions of the partial program correspond to policies for the SMDP. In the rest of this paper, we speak of policies and completions interchangeably.

[2] Assuming the passenger set is only reset after a dropoff is like assuming the taxi only communicates with the central dispatcher after serving the current passenger.

As in the MDP case, we can define the value function $V^{\pi}(\omega)$ as the expected total reward gained in the environment if we start at $\omega$ and continue the partial program, making choices using $\pi$. We are interested in finding the *hierarchically optimal* $\pi$, which must satisfy $\forall \omega, \pi' \; V^{\pi}(\omega) \geq V^{\pi'}(\omega)$. Define $Q^{\pi}(\omega, u)$ to be the expected total reward gained if we begin by making choice $u$ in $\omega$, then continue the partial program, making choices using $\pi$.[3] In this paper, we always work with undiscounted value and Q-functions, and further restrict attention to completions that eventually terminate with probability 1.

One of the virtues of hierarchical RL is that the structure in the partial program yields an additive decomposition of the Q-function. Consider the navigation choice at $\omega^{NC}$ in the trajectory shown in Figure 2. $Q(\omega^{NC}, u)$ is the expected reward over all future trajectories. This can be written as $Q_r(\omega^{NC}, u) + Q_c(\omega^{NC}, u) + Q_e(\omega^{NC}, u)$, where $Q_r$ is the expected reward while doing $u$, $Q_c$ is the expected reward after $u$, but before exiting the current subroutine, and $Q_e$ is the expected reward gained after exiting the subroutine. The motivation is that each of these components might be amenable to *state abstraction*. At $\omega^{NC}$, $Q_r$ is the reward for moving one step, which is constant, while $Q_c$ is the reward until reaching the current goal, which only depends on the taxi's position and destination. Neither of these components depends, say, on the passenger's destination [Dietterich, 2000].

There exist algorithms for learning $Q$ and for learning the three components separately [Andre and Russell, 2002]. Note that the behaviour of the *runtime agent* given a learnt Q-function is conceptually simple—at a state $\omega$, pick $u$ to maximize $Q(\omega, u)$ (in the decomposed case, just add the components together).

A problem with the 3-part decomposition is that the $Q_e$ component will often depend on many variables. The $Q_r$ and $Q_c$ components are *local*, i.e., they refer to the total rewards gained within some subroutine, whereas $Q_e$ measures all future rewards received after the current subroutine. At the navigation choice in the previous example, $Q_e$ includes the rewards gained while dropping off the current passenger and serving all future passengers. As we scale up to domains with huge numbers of variables, $Q_e$ will typically depend on most of them. Thus the number of parameters to learn will scale exponentially with the number of state variables. Our goal is to avoid having to represent and learn $Q_e$, yet preserve the possibility of hierarchically optimal behaviour.

[3] In future, we will often leave out the $\pi$ superscript when it is obvious from context.

## 4 Recursive Q-function decomposition

We begin by expressing $Q_e$ in terms of $Q_r$ and $Q_c$ functions at higher levels of the hierarchy. Given a subroutine $\sigma$, and a state $\omega$ and choice $u$ occurring in the call stack at or below $\sigma$, write $P_e^{\pi,\sigma}(\omega'|\omega, u)$ for the probability of exiting $\sigma$ at state $\omega'$ given that we choose $u$ at $\omega$, then follow $\pi$. We also write $P_e^{\pi,\sigma}(\omega'|\omega)$ to mean $P_e^{\pi,\sigma}(\omega'|\omega, \pi(\omega))$. In the special case where $\omega$ occurs in $\sigma$ itself, we will leave out the $\sigma$ superscript, and refer to this conditional distribution as the *exit distribution* of $\sigma$. Write $V_r^{\pi}(\omega) = Q_r^{\pi}(\omega, \pi(\omega))$ and similarly for $V_c$ and $V_e$ (as with $Q$, we will often omit the $\pi$ superscript of $V$ and $P_e$). Then,

$$\begin{aligned}
Q_e(\omega^{NC}, u) &= E_{P_e(\omega^{PA}|\omega^{NC},u)}[V(\omega^{PA})] \\
&= E_{P_e(\omega^{PA}|\omega^{NC},u)}[V_r(\omega^{PA}) + V_c(\omega^{PA}) + \\
&\quad V_e(\omega^{PA})] \\
&= E_{P_e(\omega^{PA}|\omega^{NC},u)}[V_r(\omega^{PA}) + V_c(\omega^{PA}) + \\
&\quad\quad E_{P_e(\omega^{PP}|\omega^{PA})}[V(\omega^{PP})]] \\
&= E_{P_e(\omega^{PA}|\omega^{NC},u)}[V_r(\omega^{PA}) + V_c(\omega^{PA}) + \\
&\quad\quad E_{P_e(\omega^{PP}|\omega^{PA})}[V_r(\omega^{PP}) + V_c(\omega^{PP}) + \\
&\quad\quad\quad E_{P_e(\omega^{SP}|\omega^{PP})}[V_r(\omega^{SP}) + V_c(\omega^{SP})]]]
\end{aligned}$$

This derivation mirrors the structure of the program in Figure 2. Note that the figure depicts one possible trajectory of the program; other trajectories may have different call structures. Nevertheless, all trajectories passing through $\omega^{NC}$ will contain (random) exit states $\omega^{PA}, \omega^{PP}$, and $\omega^{SP}$, and those are the only states referred to.[4] The following theorem generalizes the derivation.

**Theorem 1.** *Let $\omega$ be a joint state of an ALisp program with stack $(\sigma_n, \ldots, \sigma_0)$. Given a choice $u$ at $\omega$ and a completion $\pi$, define random variables $\omega^1, \ldots, \omega^n$, where $\omega^i$ is the exit state of $\sigma_i$ that is reached after choosing $u$ at $\omega$ and following $\pi$ thereafter. Then,*

$$\begin{aligned}
Q_e(\omega, u) &= E_{P_e(\omega^n|\omega,u)}\left[V_r(\omega^n) + V_c(\omega^n) + V_e(\omega^n)\right] \\
&= E_{P_e(\omega^n|\omega,u)}[V_r(\omega^n) + V_c(\omega^n) + \\
&\quad\quad E_{P_e(\omega^{n-1}|\omega^n)}[V_r(\omega^{n-1}) + V_c(\omega^{n-1}) + \\
&\quad\quad \ldots + E_{P_e(\omega^1|\omega^2)}[V_r(\omega^1) + V_c(\omega^1)]
\end{aligned}$$

Since $V_r$ and $V_c$ can be computed from $Q_r$ and $Q_c$, this theorem shows that knowledge of $Q_r$, $Q_c$, and the exit distributions is enough to act optimally.

It might seem like we have exchanged one problem for another, since instead of $Q_e$, we now have to learn exit probability distributions $P_e$. But in fact, the individual exit distributions often have a lot of structure, such as determinism and conditional or context-sensitive independencies, which might be known beforehand. For example, at the navigation choice $\omega^{NC}$ in Figure 2, $P_e(\omega^{PA}|\omega^{NC})$ deterministically leads to the state where the taxi is at the passenger location, and $P_e(\omega^{PP}|\omega^{PA})$ deterministically leads to the state where the passenger is now in the taxi. Also, $P_e(\omega^{SP}|\omega^{PP})$ either results in termination, or keeps the taxi at its current location and reinitializes each passenger independently, and so the only quantities that need to be estimated are the termination probability and the distribution over an individual passenger's source, destination, and generosity.

[4] We are making use of the guaranteed termination of legal ALisp programs.

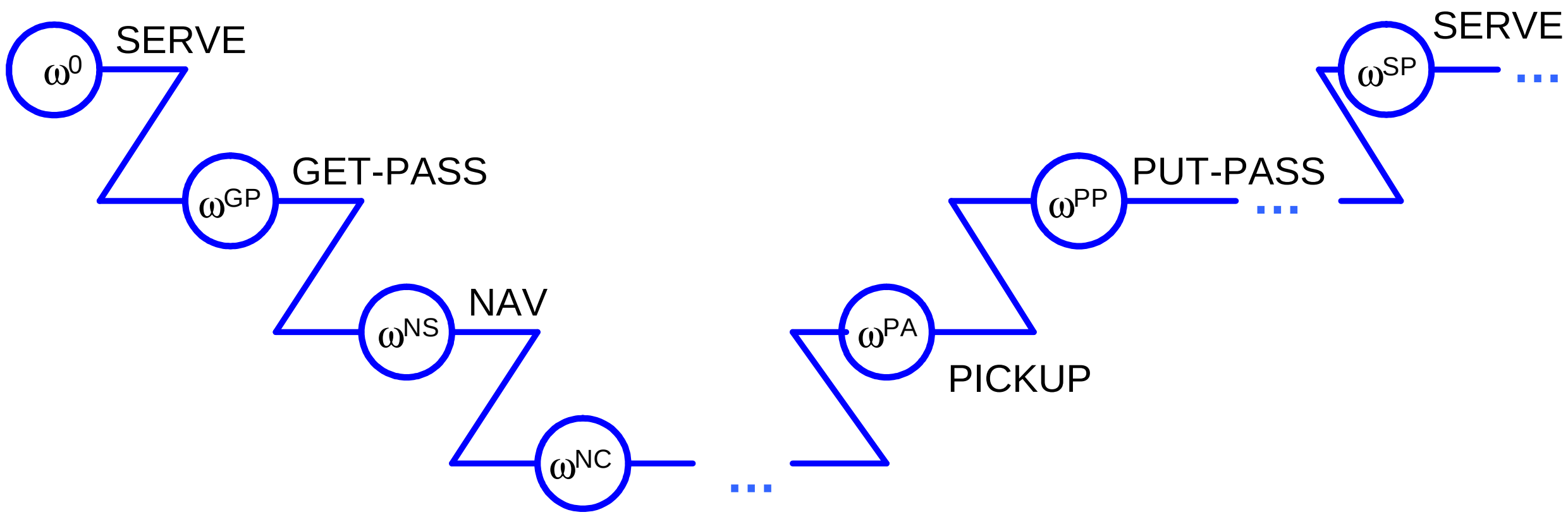

Figure 2: An execution trajectory of the partial program in Example 1. The circles represent successive joint states. The vertical position of a circle depends on the depth of the stack of choices/calls at that state. With the exception of the leftmost state $\omega^0$, which is at the call statement in `taxi-main`, the circles are labelled with label at the corresponding choice or action statement. For example, $\omega^{GP}$ is a joint state where the program counter points to the choice statement with label `GP`.

### 4.1 Computing $Q_e$ at runtime

**Algorithm 1** Functions for computing $Q_e$ at runtime. Initialize is called by the overall ALisp interpreter when starting to run the program. CallSub is called when a subroutine is called. `QStack` is a global stack maintained by the interpreter. The stack is popped when leaving a subroutine. Decisions are made by maximizing the Q-function at the top of the stack. The operations `Expectation, Add`, and `GetValueFn` are assumed provided as blackboxes.

**function** INITIALIZE
  $Q \leftarrow Q_r + Q_c$ of top level
  PUSH($Q$, QStack)
**end function**

**function** CALLSUB($\sigma$)
  $Q_{\text{par}} \leftarrow$ TOP(QStack)
  $V_e \leftarrow$ GETVALUEFN($Q_{\text{par}}$)
  $Q_e \leftarrow$ EXPECTATION($V_e, P_e, \sigma$)
  $Q \leftarrow$ ADD($Q_r, Q_c, Q_e$)
  PUSH(Q, QStack)
**end function**

Given $Q_r$, $Q_c$ and $P_e$, we need to compute $Q$ from them to act optimally at runtime. The recursive nature of Theorem 1 suggests that the entire computation need not be repeated for each state, but can happen whenever a call is made, as shown in Algorithm 1.

The algorithm maintains a stack of Q-functions parallel to the call stack. Of course, there is really only one Q-function, but the point is that the Q-function on the stack for a particular subroutine can be a compact representation that is valid only for states encountered in that subroutine. Moreover, the operations within the algorithm, such as `Expectation`, can operate directly on these representations. For example, we might use decision trees for Q-functions, DBNs for the exit distributions, and the methods of [Boutilier *et al.*, 1999] to operate on them efficiently.

When calling a subroutine, the exit value function for that subroutine is computed using the parent's Q-function. This is then combined with the child's exit distribution to get a representation of the Q-function for the child, which is added to the stack. Decisions within the subroutine are made by maximizing this Q-function. An intuitive description of the overall procedure is that each subroutine "passes in" its exit value function to children, who compute their Q-function from it.

### 4.2 A learning algorithm for $P_e$

Prior work [Andre and Russell, 2002] has described the HORDQ algorithm, which learns $Q_r$, $Q_c$, and $Q_e$ based on samples from a GLIE policy. Now $P_e$ satisfies the Bellman equation

$$P_e^\pi(\tilde{\omega}|\omega, u) = P(\tilde{\omega}|\omega, u) + \sum_{\omega' \in SS} P(\omega'|\omega, u) P_e^\pi(\tilde{\omega}|\omega', \pi)$$

where $P(\cdot|\omega, u)$ is the distribution over the next choice state at the same level as $\omega$ encountered after choosing $u$ in $\omega$, and $SS(\omega)$ is the set of choice states in the same subroutine as $\omega$. We can therefore modify the original HORDQ algorithm to a Hierarchically Optimal Cascaded Q-Learning algorithm (HOCQ) that learns $P_e$ instead of $Q_e$. Every time we encounter successive states at the same level $\omega$ and $\omega'$ (with possibly some intervening states at lower levels), we set the new value of the conditional distribution $P_e(\tilde{\omega}|\omega, u)$ to

$$(1 - \eta)P_e(\tilde{\omega}|\omega, u) + \eta P(\omega'|\omega, u)P_e(\tilde{\omega}|\omega', \pi(\omega'))$$

where $\eta$ is a learning rate parameter. When we encounter a state $\omega$ that is followed by a state $\omega'$ at the parent level, we set $P_e(\tilde{\omega}|\omega, u)$ to

$$(1 - \eta)P_e(\tilde{\omega}|\omega, u) + \eta\delta_{\omega'}(\tilde{\omega})$$

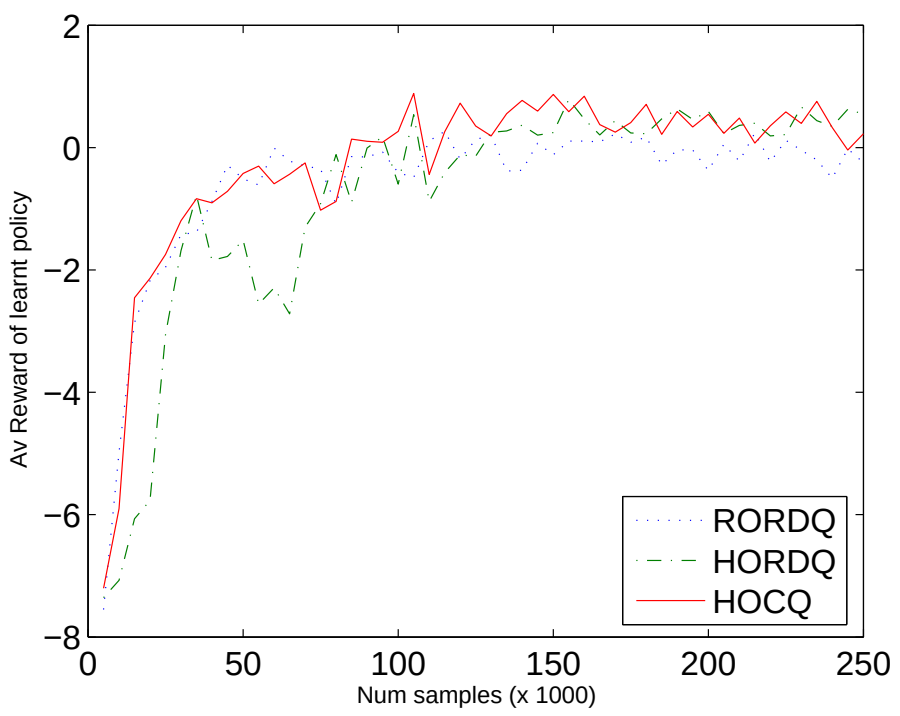

Figure 3: Learning curves for RORDQ, HORDQ, and HOCQ in Example 1, in a 6 by 6 world with two passengers (15552 states). No state abstraction was used for $Q_e$ or $P_e$. All curves averaged over 10 runs.

Figure 3 compares the performance of HOCQ with HORDQ and its recursively optimal counterpart RORDQ on Example 1. Somewhat surprisingly, even without state abstraction of $Q_e$ or $P_e$, HOCQ has a slight edge over HORDQ. Note also that RORDQ converges to a slightly suboptimal policy (essentially, it is too willing to serve generous passengers travelling to remote locations).

## 5 Additive irrelevance

The previous section showed how to replace $Q_e$ with the more "local" quantity $P_e$, but we are still not making full use of this locality. $P_e$ describes the exit distribution of all the state variables, even those that have no connection with what is going on in the subroutine. This seems unnecessary. For example, a stock-owning taxi driver shouldn't have to condition her low-level navigation decisions on unrelated variables such as the current state of the NASDAQ exchange, even if this variable does affect the exit value. In this section, we formalize such intuitions. We begin with an obvious but useful fact.

**Lemma 1.** *(Additive Irrelevance) If* $Q(\omega, u) = Q_1(\omega, u) + Q_2(\omega)$*, then* $\forall\omega \ \arg\max_u Q(\omega, u) = \arg\max_u Q_1(\omega, u)$.

A consequence of this fact [Andre, 2003] is that if the exit distribution of the current subroutine at a state $\omega$ is unique, i.e., it does not depend on the choice made at $\omega$, then $\arg\max_u Q(\omega, u) = \arg\max_u Q_r(\omega, u) + Q_c(\omega, u)$. This condition applies at the navigation choice, since given any joint state $\omega$ arising in the `nav` subroutine, there is only one possible exit state, namely the one in which the taxi is at its navigation goal, and all other variables are unchanged. Thus, there is no need to even compute $Q_e$ at this state. On the other hand, the condition does not hold at the passenger choice $\omega^{GP}$, since the choice of passenger affects the taxi position after the dropoff.

In practice, the unique exit condition will rarely hold. Even at the navigation choice, adding a fuel variable would destroy it, since the navigation choices would affect the amount of fuel when exiting. We thus look for a more general way to apply Lemma 1. Here is a motivating example.

**Example 2.** Consider modifying Example 1 by adding a high-dimensional variable $\omega_g$ that represents the state of all traffic on the roads. $\omega_g$ evolves according to a transition distribution $P(\omega'_g|\omega_g)$, independently of the rest of the state, each time a passenger is served. Further, whenever some statistic of $\omega_g$ exceeds a threshold, a constant toll is charged per passenger (after a successful `dropoff`).

Since the unique exit condition does not hold at $\omega^{GP}$, it would seem that we must learn $P_e(\omega^{SP}|\omega^{GP}, u)$, which includes the exit distribution of `g`. Note, however, that $u$ does not actually affect the exit distribution of `g`. Also, a bit of thought shows that the exit value function can be decomposed as $V(\omega^{SP}) = V_1(\omega^{SP}_{x,y,pass}) + V_2(\omega^{SP}_g)$, where `pass` is the source, destination and generosity of all the passengers, $V_2$ consists of the expected future tolls, and $V_1$ includes all the other rewards. Since $V_2$ only depends on `g`, its contribution to $Q_e$ will be constant with respect to $u$, and may therefore be ignored. So we just need to compute the expected exit value of $V_1$, which only requires knowing $P_e(\omega^{SP}_{x,y,pass}|\omega^{GP})$. This argument can be generalized.

**Definition 1.** Given a transition distribution $P(\omega'|\omega, u)$, we say that a set of variables $D$ is *decoupled from the choice at* $\omega$ *with respect to* $P$ if $\omega'_D$ is conditionally independent of $u$ given $\omega$ under the distribution $P$.[5] We call a variable (or set) of variables *coupled* if it is not a member (or subset) of any decoupled set.

We will typically take $P$ to be the exit distribution. In Example 2, $\{$`g`$\}$ is decoupled from the choice at $\omega^{GP}$ with respect to $P_e$. To incorporate additive irrelevance into the recursive decomposition, we need a definition that applies to entire subroutines.

**Definition 2.** A set of variables $D$ is *decoupled from subroutine* $\sigma$ if, for any $\omega$ occurring in $\sigma$ or its descendants, and any completion $\pi$ within $\sigma$, $D$ is decoupled from the choice at $\omega$ with respect to $P^\pi_e$.

**Definition 3.** A triple $(D, R, V_1)$ satisfies the *factored exit condition* with respect to a subroutine $\sigma$ and function $V$ if $D$ is decoupled from $\sigma$ and, for every exit state $\omega'$ of $\sigma$, $V(\omega') - V_1(\omega'_R)$ only depends on $\omega'_D$.

[5] Strictly speaking, $u$ is not a random variable, so this just means that $P(\omega'_D|\omega, u)$ is the same for each $u$.

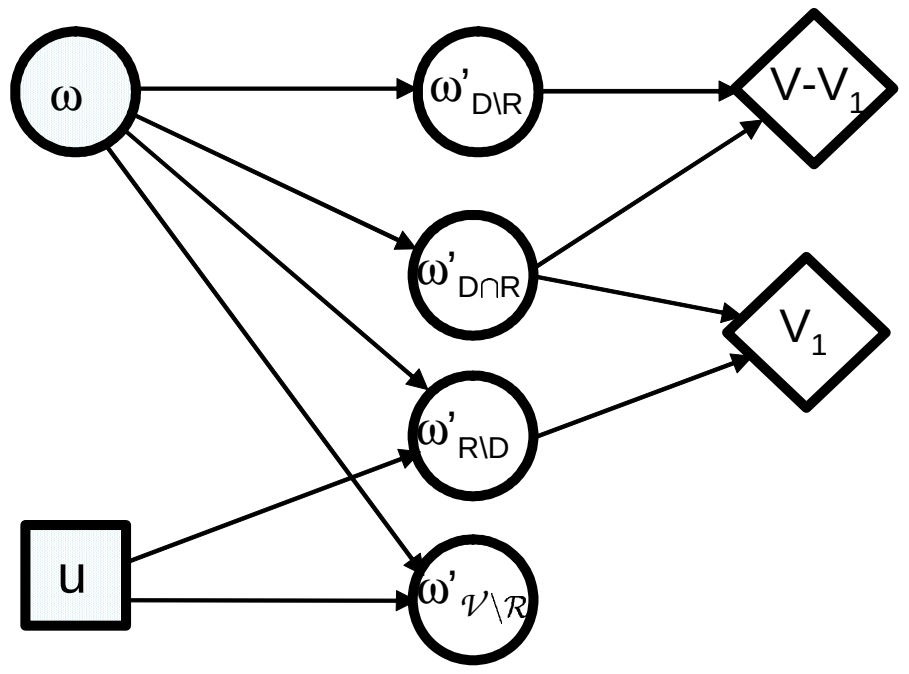

Figure 4: The factored exit condition. This condition must hold all states occurring in a subroutine $\sigma$ and its descendants, with respect to the exit distribution $P_e^\sigma$ of $\sigma$. $\mathcal{V}$ is the set of all variables.

Figure 4 illustrates this condition. We call $V_1$ the *reduced value function* and $R$ the set of its *relevant variables*. For the passenger choice in Example 2, we can satisfy this condition using decoupled variables $\{$`g`$\}$, relevant variables $\{$`x,y, pass`$\}$, and the reduced value function that doesn't include tolls.

**Theorem 2.** *Suppose $(D, R, V_1)$ satisfies the factored exit condition with respect to $\sigma$. Let $\omega$ be a joint state with stack $(\sigma_0, \ldots, \sigma_{m-1}, \sigma, \sigma_{m+1}, \ldots, \sigma_n)$. Let the random variables $\omega^1, \ldots, \omega^n$ denote the exit states of these subroutines. Then,*

$$\begin{aligned} Q_e(\omega, u) \quad = \quad & E_{P_e(\cdot|\omega,u)}[V_r(\omega^n) + V_c(\omega^n) + \\ & E_{P_e(\cdot|\omega^n)}[V_r(\omega^{n-1}) + V_c(\omega^{n-1}) + \\ & \ldots + E_{P_e(\cdot|\omega^{m+1})}[V_1(\omega_R^m)]]] + f(\omega) \end{aligned}$$

*where $f$ does not depend on $u$. In particular, optimal decisions can be made by maximizing $Q_e - f$.*

This theorem justifies a modification to Algorithm 1 in which each subroutine passes a reduced value function to its child. Algorithm 2 makes this precise. In addition to $Q_r, Q_c$, and $P_e$, the algorithm takes as input, for each subroutine, sets $D$ and $R$, and a blackbox function `Reduce` that takes in a representation of an exit value function $V$ and produces $V_1$ such that $(D, R, V_1)$ is a factored exit triple for $V$. Such a blackbox function is often easy to write if we have a structured representation of $V$. If we do not have this information for a subroutine, we can set $R$ to $\mathcal{V}$, $D$ to the empty set, and `Reduce` to the identity function.

Thus, using the factored exit condition will speed things up at runtime, since we only compute expectations of the reduced exit value function. There will also be an improvement during learning; we now only need to learn the exit distributions of the relevant variables for each subroutine, and can also make use of the factorization $P(\omega'_R|\omega, u) = P(\omega'_{R\cap D}|\omega)P(\omega'_{R\setminus D}|\omega, u)$. It is straightforward to extend the basic HOCQ algorithm of Section 4.2 to take these facts into account. The extent of the gains will depend on the number of relevant variables and their domain size, and on how many variables in the current state affect the exit values of the relevant variables.

**Algorithm 2** Modification to Algorithm 1 that takes factored exit conditions into account.

```
function CallSub(σ)
    Q_par ← Top(QStack)
    V_e ← GetValueFn(Q)
    V̄_e ← Reduce(V_e)
    Q_e ← Expectation(V̄_e, P_e, σ)
    Q ← Add(Q_r, Q_c, Q_e)
    Push(Q, QStack)
end function
```

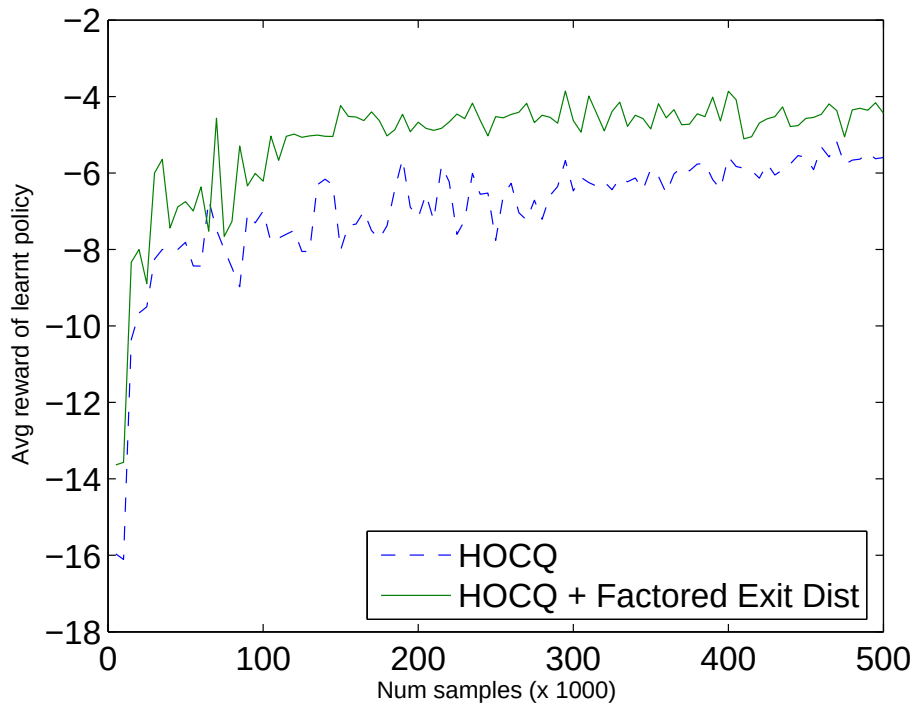

Figure 5: Learning curves for HOCQ with and without factored exit condition in Example 2, averaged over 10 runs.

In Figure 5, we measure the effect of using the factored exit condition in Example 2. The traffic variable takes 10 different values. The per-dropoff transition matrix for traffic and the toll vector were chosen randomly at the start of the experiment. Without the factored exit condition, the algorithm learns to navigate as before, but has difficulty reliably learning to choose passengers optimally. When factored exit distributions are taken into account, the algorithm learns as quickly as in Figure 3.

## 6 Separating a subroutine from its context

In some cases, the factored exit condition does not simplify the problem much. One family of such cases occurs when choices made in a subroutine affect the number of time steps that elapse before exiting the subroutine, which in turn affects the exit value of most of the state variables.

**Example 3.** Consider the environment of Example 2, modified so that the traffic variable evolves once per timestep, rather than only when a passenger is dropped off. For convenience, we also add the state variable `t`, which equals the current timestep.

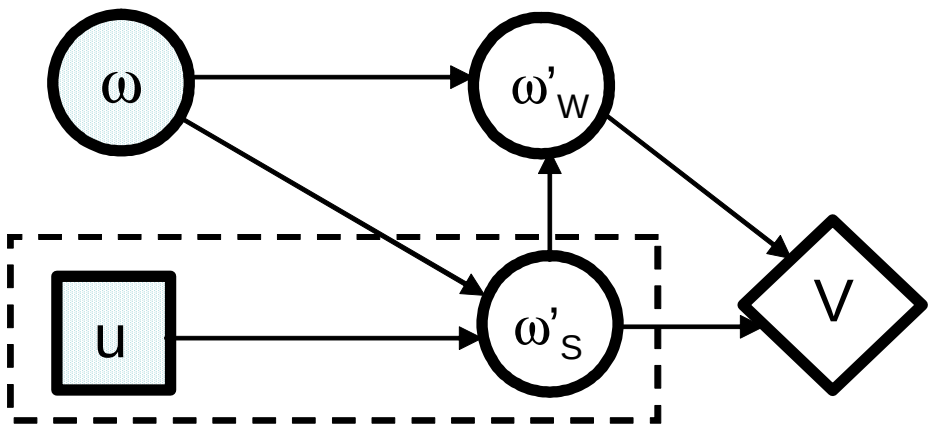

Figure 6: In this picture, $S$ is a separator for $W$. The dashed box surrounds that part of the distribution that is used at runtime by the subroutine containing $\omega$ to compute the expected value of $\tilde{V}$. The part outside the box is used by the parent to compute $\tilde{V}$.

Suppose we are at the passenger choice in Example 3 and choosing between two passengers, such that one of them will take longer to serve, but is more generous. Since the traffic evolves once per timestep, its exit value is coupled with the choice of passenger, and there is no useful application of Theorem 2.

The second type of example that causes problems is when events occurring within a subroutine can change the state in a way that affects all future rewards.

**Example 4.** Consider the environment of Example 2, modified so that some map squares are designated as "cliffs". Moving onto one of these squares results in a large negative reward and immediate termination of the episode. For correctness, the partial program is also modified to terminate when at a cliff.

Suppose we are at the passenger choice of Example 4. Let `c` be a Boolean variable that is true if the taxi has moved on to a cliff square. Then, assuming moves are noisy, `c` is coupled with our choice of passenger; for example, picking up a passenger in a region of the map with many cliff squares would carry a high risk of accidentally entering one of them. But it is no longer possible to find a nontrivial additive decomposition of the exit value function in which one of the components does not depend on `c`. Intuitively, when deciding between a stingy passenger in the plains, and a generous one in the mountains, the optimal decision depends on *how much* future expected reward is being lost in the second case due to the possibility of falling off a cliff. So we cannot ignore any portion of the exit value function.

A commonality between the last two examples is the existence of a small set of variables that makes the exit values of all the variables relevant to $Q_e$, either by making them all coupled with the current choice, as with `t` (the current timestep) in the first example, or by jointly participating with them in some value function component as with `c` in the second case. We can formalize this idea.

**Definition 4.** Given $P(\omega'|\omega, u)$, and sets of variables $S$ and $W$, we say that $S$ is a *separator* for $W$ if $\omega'_W$ is conditionally independent of $u$ given $\omega$ and $\omega'_S$.

Figure 6 illustrates this condition. At the passenger choice in Example 3, $\{$`t, x,y`$\}$ is a separator for $\{$`g,pass`$\}$. At the passenger choice in Example 4, $\{$`c, x,y`$\}$ is a separator for $\{$`c, pass, x,y`$\}$. As shown in this example, it is possible for $S$ and $W$ to overlap.

Let's look a bit more closely at Example 3. Imagine that the subroutines are separate entities, and each subroutine "knows" only about the direct effects of its actions on local rewards and on exit values of some subset of the variables. Thus the `serve` subroutine only knows how its choice of passenger affects rewards within the subroutine and the exit distribution of $\{$`t, x,y`$\}$. Given this knowledge, what additional information is needed to make optimal decisions at $\omega^{GP}$? The answer is that we need to know the "projection" of the exit value function onto the variables in the separator set $\{$`t, x,y`$\}$:

$$\tilde{V}_{\omega^{GP}}(\omega^{SP}_{t,x,y}) \stackrel{\text{def}}{=} E_{P_e}[V(\omega^{SP})|\omega^{GP}, \omega^{SP}_{t,x,y}]$$

where the right hand side makes sense independently of $u$ because $V(\omega^{SP})$ depends only on the variables in $W$, and `t, x,y` is a separator for $W$. Knowing $\tilde{V}$ is sufficient, for then we can compute

$$Q_e(\omega^{GP}, u) = E_{P_e}[\tilde{V}_{\omega^{GP}}(\omega^{SP}_{t,x,y})|\omega^{GP}, u]$$

using the known distribution $P_e(\omega^{SP}_t|\omega^{GP}, u)$. Intuitively, the parent computes and "passes in" $\tilde{V}$ to the child, who uses it, together with $P_e$ of the separator variables, to compute $Q$, as shown in Figure 6. In general,

**Definition 5.** Suppose $S$ is a separator for $W$ at $\omega$. Let $f(\omega')$ be any function on exit states $\omega'$ that only depends on $\omega'_{S\cup W}$. The *projection of $f$ onto $S$* is defined as

$$\tilde{f}_\omega(\omega'_S) = E_{P(\omega'_W|\omega,\omega'_S)}[f(\omega'_{W,S})]$$

**Lemma 2.** *Suppose $S$ is a separator for $W$ at $\omega$, and that the (possibly reduced) exit value function $V(\omega')$ depends only on the variables $W \cup S$ . Then*

$$Q_e(\omega, u) = E_{P_e(\omega'_S|\omega,u)}[\tilde{V}_\omega(\omega'_S)]$$

In light of Lemma 2, we can modify Algorithm 2 to take separators into account, as shown in Algorithm 3. The algorithm takes, as an additional input, sets $S$ and $W$ for each pair of subroutines $\sigma_1, \sigma_2$ where $\sigma_1$ can call $\sigma_2$, such that $S$ is a separator for $W$ within $\sigma_2$ and the exit value function of $\sigma_2$ only depends on $S \cup W$. The learning algorithm must also be modified to separately learn $P(\omega'_W|\omega, \omega'_S)$ and $P^\pi_e(\omega'_S|\omega, u)$.

What have we gained by this decomposition? The first component is a kind of "general knowledge", which will often be unaffected by aspects of $\omega$ that pertain to the details of the particular task being done, and can thus be learnt using samples from many different tasks. At the passenger choice in Example 3, this component equals $P(\omega'_g|\omega_g, \omega'_t)$, the distribution of how traffic evolves over time, while in

**Algorithm 3** Modification to Algorithm 2 that makes use of separator sets.

```
function CALLSUB(σ)
    Q_par ← TOP(QStack)
    V_e ← GETVALUEFN(Q)
    V̄_e ← REDUCE(V_e)
    Ṽ_e ← PROJECT(V̄_e, S)
    Q_e ← EXPECTATION(Ṽ_e, P_e, σ)
    Q ← ADD(Q_r, Q_c, Q_e)
    PUSH(Q, QStack)
end function
```

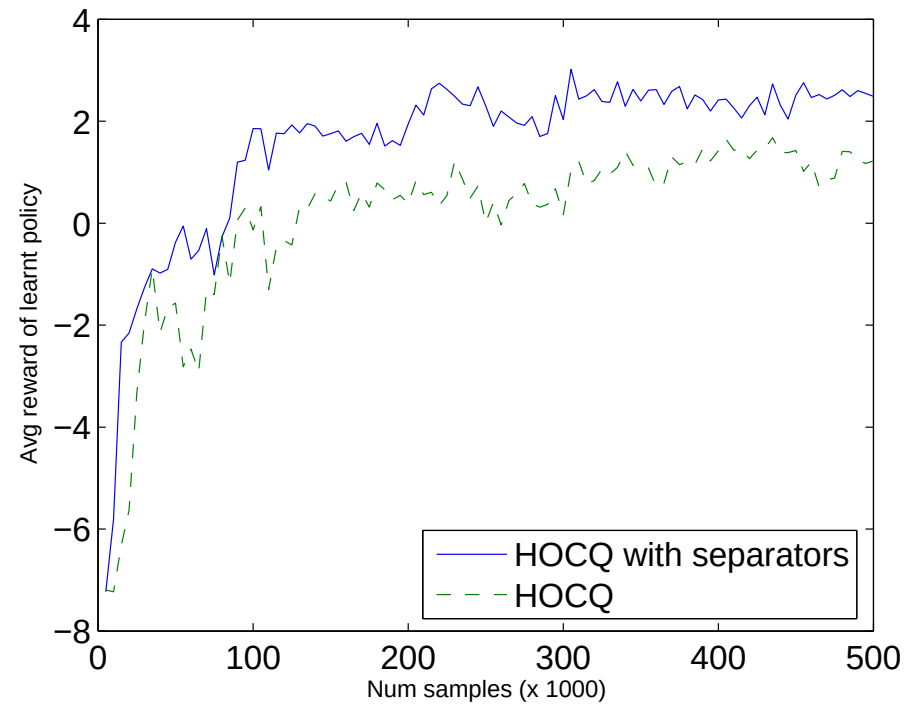

Figure 7: Learning curves with and without making use of separators

Example 4, it equals $P(\omega'_g|\omega_g)$, the distribution of how traffic evolves each time a passenger is served (note that in this case, the distribution does not actually depend on the exit values of any of the separator variables).

The second component of our decomposition is the exit distribution of just the separator variables, which are directly related to the task being performed in the subroutine. At the passenger choice in Example 3, this component is $P(\omega'_{t,x,y}|\omega, u)$, the exit distribution of time and position. Assuming deterministic moves, and a completion that follows the shortest path towards its goal, this distribution depends only on the current location and the passenger's source and destination. At the passenger choice in Example 4, the second component is the exit distribution of `c, x,y`, i.e., a distribution over taxi position and whether the taxi falls of a cliff during the subroutine. Again, this depends on the taxi's location and the passenger's source and destination.

In Figure 7, we measure the effect of using the separator decomposition in Example 3. We let the traffic variable be an integer between 0 and 10, and made it follow a random walk with a slight upward drift over time. The toll is proportional to the square of the traffic. Without the separator conditions, the algorithm must directly learn the exit distribution of traffic as a function of passenger choice, and learns quite slowly. When the separator conditions are used, the algorithm can now take advantage of the abstractions described above, and learns much faster.

# 7 Local policies

The exit distribution of the separator variables is not always as easy to abstract as in the last two examples.

**Example 5.** Consider a combination of Examples 3 and 4 where traffic evolves every timestep and there are cliffs.

Suppose we are at the navigation choice in this example, and the traffic is below the threshold beyond tolls will be charged. In a few steps, we will reach a fork in the road, and must choose between two paths. The first path is safe and slow, and the other is fast goes through mountainous regions. Suppose the rewards are such that if the traffic doesn't change much before we reach the fork, then it will make sense to choose the slow route, but if the traffic level rises sufficiently during this period, it will make sense to risk the fast route to increase the chance of avoiding the toll. The traffic when making the future choice, in turn, depends on the current traffic. So, somewhat surprisingly, the exit distribution of `t` for the optimal completion depends on the current traffic, even though traffic has no direct effect on events within the subroutine (recall that it only affects the toll, not how fast the taxi moves). The problem is that separated variables may nevertheless affect future decisions made in the subroutine, and so any quantity that is relevant to predicting future values of these variables becomes relevant to the exit distribution of the separator variables.

Unfortunately, it does not seem possible to handle such cases exactly while keeping the exit distribution compact. In this example, we could probably ignore the traffic without much harm, but in general, such an approximation would systematically underestimate the exit value, because it would be based on the assumption that future decisions are made using an unsafe abstraction. As an alternative, we show a further level of structure in the exit distribution that allows more principled approximations.

The dependence between the exit distribution of the separator variables and the current value of the separated variables is not arbitrary; it only happens via the separated variables' effect on the exit value function, and therefore on the "local policy" at future choices in the subroutine.

**Definition 6.** Let $\pi$ be a completion, $\omega_W$ an instantiation to a set $W$ of variables, and $L = \mathcal{V} \setminus W$. The *local policy* $\pi_{\omega_W}$ is defined by $\pi_{\omega_W}(\omega_L) = \pi(\omega_{W,L})$.

Suppose we are at a state $\omega^0$ in some subroutine $\sigma$, and that $D$ is decoupled from the root subroutine of the program. Let the trajectory of states after $\omega^0$ be $\omega^1, \omega^2, \ldots$. There is a corresponding random sequence of local policies $\pi^i = \pi_{\omega^i_D}$. Let $\vec{\pi} = (\pi^1, \pi^2, \ldots)$ and note that $\vec{\pi}$ is independent of $u$ given $\omega$. We now rewrite

$$P_e(\omega'_S|\omega, u) = \sum_{\vec{\pi}} P(\vec{\pi}|\omega) P(\omega'_S|\omega, u, \vec{\pi}) \qquad (1)$$

where $P(\vec{\pi}|\omega)$ is once again "general knowledge", and conditional on a given $\vec{\pi}$, future decisions in the subroutine will no longer depend on variables in $D$.

In our quest for simplicity, we have replaced a large but finite, observable object ($\omega_D$) with an unbounded, unobservable one ($\vec{\pi}$). This might not seem like progress. But in fact, we never have to represent $\vec{\pi}$ explicitly, since it only serves as a hidden mixture component. Also, we claim that (1) exposes numerical structure for function approximators and provides a useful "hook" for specifying prior knowledge. The infinite mixture can often be well approximated by a small finite one, where each of the components in the reduced model corresponds to an event such as "I'm not in a hurry now, but will be when I reach the upcoming fork".

## 8 Related Work and Conclusions

In addition to the work mentioned in the introduction, many other bodies of research have explored the ideas of the decomposition of value functions and exit distributions in hierarchical reinforcement learning. [Dean and Lin, 1995], [Hauskrecht *et al.*, 1998], and [Parr, 1998] are all based on the idea of dividing the state space of an MDP into regions which communicate via small sets of interface states (which would be subroutine call and exit states, in our case), solving the subproblems separately, and combining the solutions somehow. These papers mainly utilize structure in the state-transition graph, and deal with planning given a model. In contrast, we have focused on viewing states as factored into a set of variables, and on how structure within this factored representation can lead to representational and statistical benefits in learning. An interesting question is whether this factored structure also has algorithmic benefits, i.e., can it be used in conjunction with the methods of the aforementioned papers to yield more efficient planning algorithms? We note also that our notion of local policies is reminiscent of the policy-cache idea of [Parr, 1998], and it might be possible to use the methods for policy-cache construction therein to approximate the mixture representation of the exit distribution in Section 7.

A recent paper [Meuleau *et al.*, 2006] studies the oversubscription planning problem, in which we are given a set of tasks which consume resources and must complete as many of them as possible given global resource constraints. The framework can be viewed as an ALisp program with a top-level loop that chooses the next task to do and, for each task, a subroutine that chooses between actions for that task and aborting. The state is factored into global variables (e.g. resource amounts), variables local to each task and, for each task, an indicator for whether that task is completed. In our terminology, while engaged in a particular task, the exit values of the resource amounts and the completion indicator for the current task are separators for the remaining variables. The paper shows that under a reset assumption, which states that if a task is aborted before completion, its local variables revert to the initial state, a dynamic programming algorithm that alternates between planning at the top-level and planning for subtasks is guaranteed to converge. It would be interesting to see if the algorithm can be extended to general hierarchies, and without the reset assumption. A similar algorithm is also suggested in [Dietterich, 2000].

In the options framework [Precup and Sutton, 1998], models of the expected behavior of options are created that are somewhat similar to the notion of $P_e$ in our work. A difference is that work in options thus far has usually assumed that the low-level options are specified beforehand, whereas we are interested in the problem of simultaneously planning at the low and high levels.

The primary contributions of this paper are to present a method and a set of conditions under which hierarchically optimal solutions can be obtained without large representational and parameter learning costs. Unlike previous work that reasons explicitly about exit states of subroutines, our method generalizes easily to arbitrarily deep hierarchies and captures the notion that the parent passes to the child subroutine an exit value function that is as concise as possible.

To keep the presentation focused, we have only discussed exact decomposition of the Q-function in this paper. In practice, approximate versions of the factored exit and separator conditions will be required, and the components of the decompositions will themselves need to be approximated. Applying our algorithms to larger domains will requiring facing these issues.

This work is but a step toward a full understanding of the structure of value functions. The link between this structure, the behavioural hierarchy, and the complexity of decision making needs to be investigated further.